\documentclass[authoryear,preprint,review,12pt]{elsarticle}

\usepackage{amssymb}
\usepackage{multirow}
\usepackage{amsmath}
\usepackage[colorlinks=true, linkcolor=black, citecolor=blue, urlcolor=blue]{hyperref}
\usepackage{booktabs}
\usepackage{fontspec}
\usepackage{polyglossia}
\setdefaultlanguage{english}
\setotherlanguage{bengali}

\newfontfamily\bengalifont[Script=Bengali,Language=Bengali]{NotoSerifBengali-VariableFont.ttf}

\usepackage[most]{tcolorbox} 
\tcbuselibrary{listings}     

\journal{Array}

\begin{document}

\begin{frontmatter}

\title{BanglaMedQA and BanglaMMedBench: Evaluating Retrieval-Augmented Generation Strategies for Bangla Biomedical Question Answering}

\author[iut, equal]{Sadia Sultana}
\ead{sadiasultana20@iut-dhaka.edu}

\author[iut, equal]{Saiyma Sittul Muna}
\ead{saiymasittul@iut-dhaka.edu}

\author[iut, equal]{Mosammat Zannatul Samarukh}
\ead{zannatul2@iut-dhaka.edu}

\author[iut]{Ajwad Abrar\corref{cor1}} 
\ead{ajwadabrar@iut-dhaka.edu}

\author[iut]{Tareque Mohmud Chowdhury}
\ead{tareque@iut-dhaka.edu}

\fntext[equal]{These authors contributed equally to this work.}
\cortext[cor1]{Corresponding author.}


\affiliation[iut]{organization={Department of Computer Science and Engineering, Islamic University of Technology}, 
            city={Gazipur}, 
            state={Dhaka},
            postcode={1704},
            country={Bangladesh}}


\begin{abstract}
Developing accurate biomedical Question Answering (QA) systems in low-resource languages remains a major challenge, limiting equitable access to reliable medical knowledge. This paper introduces BanglaMedQA and BanglaMMedBench, the first large-scale Bangla biomedical Multiple Choice Question (MCQ) datasets designed to evaluate reasoning and retrieval in medical artificial intelligence (AI). The study applies and benchmarks several Retrieval-Augmented Generation (RAG) strategies, including Traditional, Zero-Shot Fallback, Agentic, Iterative Feedback, and Aggregate RAG, combining textbook-based and web retrieval with generative reasoning to improve factual accuracy. A key novelty lies in integrating a Bangla medical textbook corpus through Optical Character Recognition (OCR) and implementing an Agentic RAG pipeline that dynamically selects between retrieval and reasoning strategies. Experimental results show that the Agentic RAG achieved the highest accuracy (89.54\%) with openai/gpt-oss-120b, outperforming other configurations and demonstrating superior rationale quality. These findings highlight the potential of RAG-based methods to enhance the reliability and accessibility of Bangla medical QA, establishing a foundation for future research in multilingual medical artificial intelligence.
\end{abstract}







\begin{keyword}
Retrieval-Augmented Generation (RAG) \sep Bangla Biomedical Question Answering \sep Natural Language Processing \sep Medical Multiple Choice Questions (MCQs) \sep Dataset Benchmarking and Evaluation

\end{keyword}

\end{frontmatter}



\section{Introduction}\label{sec1}

The advancement of Large Language Models (LLMs) such as GPT and LLaMA has transformed Natural Language Processing (NLP) tasks, including medical Question Answering (QA). These models excel at understanding and generating human-like text, allowing users to ask complex medical questions in everyday language. However, LLMs often produce inaccurate or hallucinated answers, particularly in medical contexts where accuracy is vital \citep{Qin2025MultilingualLLM,Qiu2024MultilingualMedicine}. This limitation arises because these models rely on fixed training data and lack direct access to external, up-to-date knowledge sources.

Retrieval-Augmented Generation (RAG) addresses this limitation by combining information retrieval with generative reasoning to improve answer accuracy \citep{lewis2020rag,zhang2024medicalrag}. By fetching relevant documents before generating answers, RAG provides responses that are better grounded in facts, particularly in medical and scientific domains. While RAG has demonstrated strong performance in English biomedical QA systems, its effectiveness in low-resource languages like Bangla, spoken by over 230 million people, remains largely unexplored \citep{Khan2023Nervous,Shafayat2024BenQA,Kaggle2023BengaliMedical}. The lack of high-quality Bangla biomedical datasets and evaluation tools limits the development of reliable QA systems, creating barriers in education and healthcare support \citep{Qin2025MultilingualLLM}.  \\

Building reliable Bangla medical QA systems involves several challenges. Large-scale Bangla biomedical QA datasets with detailed rationales do not exist, which makes model training difficult \citep{Shafayat2024BenQA,Kaggle2023BengaliMedical}. Translating medical content often distorts meaning and reduces clarity \citep{pal2022medmcqa}. Retrieval is also challenging due to sparse Bangla resources \citep{lewis2020rag,zhang2024medicalrag}. Finally, evaluating both answer correctness and rationale quality requires specialized metrics and human assessment \citep{tsatsaronis2015bioasq}. Addressing these challenges is essential to expand medical Artificial Intelligence (AI) to underrepresented languages. 

\begin{figure}[t]
    \centering
    \includegraphics[width=0.7\columnwidth]{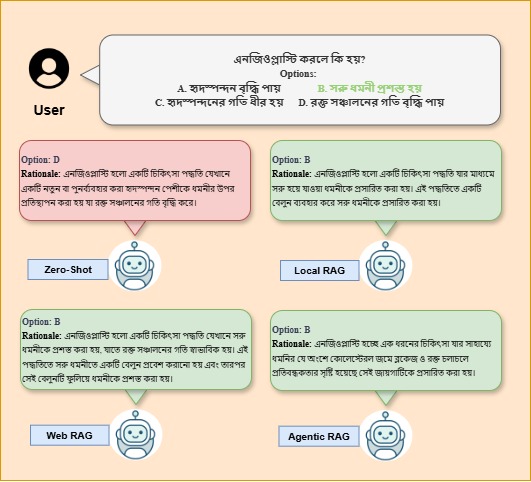} 
    \caption{The responses of llama-3.3-70b-versatile to a question from BanglaMedQA following different strategies.}
    \label{fig:fig1}
\end{figure}

To address these issues, this study develops Bangla biomedical QA datasets and systematically evaluates RAG strategies (\autoref{fig:fig1}). The contributions of this work include:  

\begin{enumerate}
    
    \item We developed BanglaMedQA and BanglaMMedBench, two publicly available Bangla Multiple Choice Question (MCQ) datasets for the medical domain, totaling 2,000 questions.\footnote{\url{https://huggingface.co/datasets/ajwad-abrar/BanglaMedQA}} BanglaMedQA contains 1,000 questions from authentic medical admission tests in Bangladesh, each accompanied by reasoning for the correct answer option. BanglaMMedBench includes another 1,000 situational and complex questions, translated and refined from the English MMedBench dataset for Bangla evaluation.

    \item We implemented and benchmarked multiple RAG variants, including Traditional RAG, Zero-Shot Fallback, Agentic RAG, Iterative Feedback RAG, and Aggregate k-values RAG. We analyzed model performance, rationale quality, and the effects of model size and retrieval strategy in a low-resource Bangla setting.
    
    \item For retrieval, we curated an external knowledge base for BanglaMedQA using Optical Character Recognition (OCR) on a standard higher secondary-level Bangla medical textbook used in Bangladesh.
 
\end{enumerate}

These contributions establish a benchmark for Bangla biomedical QA, evaluate the practical effectiveness of retrieval-augmented strategies, and advance the development of trustworthy medical Artificial Intelligence (AI) for Bangla-speaking communities. In the following sections, we present the related works, describe our proposed methodology, discuss the experimental results, and conclude with key findings and future directions.

\section{Related Works}\label{sec2}

\subsection{Medical Question Answering (QA)}
Medical Question Answering (QA) systems enable clinicians and patients to access reliable information as medical data grows rapidly. Large Language Models (LLMs) like GPT and LLaMA demonstrate strong performance in Natural Language Processing (NLP) tasks \citep{Qin2025MultilingualLLM,Qiu2024MultilingualMedicine}, but biomedical QA remains challenging due to complex terminology and limited domain-specific training. Biomedical QA tasks typically follow two formats:
\begin{enumerate}
    \item Multiple-Choice Question Answering (MCQ), exemplified by datasets such as \textit{MedMCQA} \cite{pal2022medmcqa} and \textit{PubMedQA} \cite{jin2019pubmedqa}, which test structured exam-style reasoning.
    \item Generative Answering, as seen in benchmarks like \textit{BioASQ} \cite{tsatsaronis2015bioasq}, which evaluate deeper reasoning but are more prone to inaccurate or made-up responses.
\end{enumerate}
These formats highlight the need for robust systems capable of handling diverse medical queries with high accuracy.

\subsection{Low-Resource Languages in Medical NLP}
While English benefits from extensive biomedical datasets, low-resource languages like Bangla face significant resource scarcity, limiting the development of effective QA systems \citep{11022034}. Efforts such as \textit{BEnQA} \citep{Shafayat2024BenQA}, a bilingual science QA benchmark, and Bangla health Named Entity Recognition (NER) \citep{Khan2023Nervous} provide some foundation but lack comprehensive coverage for biomedical QA reasoning. This scarcity underscores the challenge of adapting advanced QA techniques to languages with limited data, particularly in specialized domains like medicine, where accurate and contextually relevant responses are critical.

\subsection{Retrieval-Augmented Generation (RAG)}
To improve factual accuracy in QA, Retrieval-Augmented Generation (RAG) integrates information retrieval with text generation to ground responses in external knowledge \citep{lewis2020rag}. Studies show that retrieval-based models outperform purely parametric ones, enhancing factuality in biomedical QA \citep{izacard2021leveraging,zhang2024medicalrag}. However, standard RAG pipelines face issues such as retrieval errors, irrelevant passages, and unstable reasoning. To address these, several RAG variants have been proposed:
\begin{enumerate}
    \item \textbf{RAG with Fallback}: Uses Zero-Shot or parametric generation when retrieval fails \citep{lewis2020rag}.
    \item \textbf{Agentic RAG}: Allows models to decide when to retrieve or reflect, as in \textit{ReAct} \cite{yao2022react} and \textit{Reflexion} \citep{shinn2023reflexion}.
    \item \textbf{Iterative Feedback RAG}: Refines retrieval through multiple feedback loops \citep{shinn2023reflexion}.
    \item \textbf{Aggregate k-Values RAG}: Combines evidence from multiple retrieved passages for robustness \citep{zhang2024medicalrag}.
\end{enumerate}
These advancements highlight the potential of RAG to improve QA systems, particularly in challenging domains.

\subsection{Research Gap and Motivation}
Despite notable progress in Natural Language Processing (NLP) and Large Language Models (LLMs), several gaps remain in the domain of biomedical Question Answering (QA), particularly in multilingual and resource-constrained contexts.  

\begin{enumerate}
    \item \textbf{Limited Biomedical QA in Bangla and Low-Resource Languages:}
    Most existing biomedical QA systems and datasets are developed for English. Low-resource languages like Bangla remain underrepresented, limiting access for a large portion of practitioners, students, and patients in non-English speaking regions.  

    \item \textbf{Scarcity of Domain-Specific Datasets:} 
    Widely used datasets (e.g., MedQA \citep{yang2024llmmedqa}, PubMedQA, MedMCQA) are predominantly in English and often based on Western curricula. There is a lack of curated, high-quality biomedical MCQ datasets tailored to South Asian or Bangla medical textbooks, creating a significant barrier for localized applications.  

    \item \textbf{Challenges in Applying General LLMs to Biomedical MCQ Tasks:} 
    Open-domain LLMs (e.g., GPT, LLaMA, Mistral) show strong general reasoning but often lack specialized biomedical knowledge \citep{jahan2023biollm}. Without domain adaptation, these models struggle with terminologies, abbreviations, and reasoning required for medical MCQs.  

    \item \textbf{Limitations of Current Retrieval-Augmented Generation (RAG) Approaches:}
    While RAG has improved factual accuracy by grounding LLMs in external knowledge, most biomedical RAG studies focus on research articles and clinical notes, not structured MCQs. Existing RAG pipelines do not adequately handle multilingual retrieval (Bangla--English mix) or exam-style MCQ reasoning with rationales.  

    \item \textbf{Need for Explainability in Biomedical Education:}
    Current biomedical QA models mostly generate short answers without providing reasoning steps or rationales \citep{pmc11922739}. For educational and clinical training purposes, explanations are essential, yet underexplored in multilingual biomedical MCQ systems.  
\end{enumerate}

This research therefore addresses the above gaps by curating a Bangla biomedical MCQ dataset from textbooks and exam sources, implementing RAG pipelines that handle multilingual retrieval (both Bangla and English) for answer and rationale generation, enhancing both accuracy and explainability. 

\section{Proposed Methodology}\label{sec3}

This section presents the methodology used to address identified gaps in biomedical question answering. We introduce two Bangla medical Question Answering (QA) datasets - BanglaMedQA and BanglaMMedBench, along with the Higher Secondary level biology textbook corpus, written in Bangla, used as the retrieval source. The following subsection describes the implementation of multiple Retrieval-Augmented Generation (RAG) strategies, including Traditional RAG, Zero-shot fallback RAG, Agentic RAG, Iterative feedback RAG, and Aggregate retrieval RAG variants. Finally, the section concludes with the evaluation of each of the datasets to analyze accuracy, reasoning quality, and the role of web-based retrieval.

\subsection{Datasets}

To support research in medical question-answering for Bangla, we introduce two complementary datasets. BanglaMedQA is the first curated Bangla medical MCQ dataset covering foundational topics, carefully filtered for quality and consistency. Bangla MMedBench extends an English biomedical benchmark into Bangla through careful translation and pre-processing, providing reasoning-intensive questions with rationales. Together, these datasets establish both surface-level and advanced resources, serving as foundational benchmarks for evaluating Retrieval-Augmented Generation (RAG) and other Question Answering (QA) systems in a low-resource language.

\subsubsection{BanglaMedQA- Bangla Medical Corpus}
The BanglaMedQA dataset is a novel contribution, representing the first curated medical question–answer dataset in Bangla. It was compiled from authentic sources, specifically the admission tests of Bangladesh’s Medical (MBBS), Dental (BDS), and Armed Forces Medical College (AFMC) programs, spanning exam years from 1990 to 2024, covering 34 years. The dataset contains 1,000 Multiple Choice Questions (MCQs), each with four answer options, a correct answer, and an explanation for the correct answer. The questions primarily focus on surface-level factual knowledge typically required in medical admission tests.

\textbf{Quality Assurance and Data Filtering:} To ensure that the dataset is accurate, consistent, and reliable, several rigorous curation policies were followed. Initially, a total of 1,200 questions were collected from authentic medical admission test sources before applying the following filtering steps:

\begin{enumerate}
    
    \item \textbf{Exclusion of ambiguous questions:} Questions containing multiple correct answers were removed to eliminate interpretive ambiguity. During verification, five such items were identified and excluded to preserve the dataset’s clarity and validity.

    \item \textbf{Standardization of option formats:} To ensure structural consistency, questions with irregular option labels (e.g., Bangla letters such as \textnormal{\bengali{ক, খ, গ, ঘ}} or numeric formats like 1, 2, 3, 4) were discarded. Eleven such items were removed to maintain a uniform multiple-choice structure using standard English labels (A, B, C, D).

    \item \textbf{Deduplication of repeated items:} Since the data spanned multiple admission test years, duplicate questions were identified and removed to retain only unique entries. Out of the initial 1,200 collected items, 184 duplicates were eliminated, ensuring a diverse and non-redundant question set.
\end{enumerate}

By addressing these considerations, BanglaMedQA establishes a high-quality, domain-specific benchmark for evaluating question-answering systems in Bangla. The dataset not only facilitates the exploration of RAG in a low-resource language but also provides a foundational resource for future research in medical Artificial Intelligence (AI) applications in Bangladesh and the broader Bangla-speaking community.

\subsubsection{BanglaMMedBench: Translated Biomedical Corpus}

The BanglaMMedBench dataset was developed to address the absence of Bangla medical question sets involving clinical or situational reasoning. In existing Bangla medical examinations, such as MBBS or BDS admission tests, questions primarily focus on factual recall rather than scenario-based understanding. To fill this gap, we extended the English medical benchmark introduced by Qiu et al. \citep{qiu2024towards} into Bangla. The original dataset comprises situation-based medical questions inspired by the United States Medical Licensing Examination (USMLE), with accompanying rationales generated using ChatGPT. 

To translate this benchmark into Bangla, we employed the Gemini-1.5-Flash model \citep{georgiev2024gemini15} through a batch translation process. Before finalizing this choice, we conducted a comparative evaluation of several LLM-based translation methods, including Google Translate, Mistral Saba, LLaMA 3 70B, Gemini, and ChatGPT. A set of 50 sample questions was manually reviewed by a medical expert to assess translation quality, focusing on domain-specific terminology accuracy and grammatical correctness.

The evaluation revealed that Mistral Saba struggled with medical terminology, while LLaMA 3 70B frequently misinterpreted multiple-choice options. Google Translate, despite its efficiency, introduced frequent spelling and syntactic inconsistencies in Bangla. In contrast, Gemini and ChatGPT preserved medical terminology and maintained grammatical fidelity. Based on these results, computational efficiency, and expert feedback, Gemini-1.5-Flash was chosen as the final translation model.

This approach enabled the complete translation of the dataset into Bangla while preserving the original structure of questions, options, correct answers, and rationales.

After translation, several preprocessing steps were applied to ensure quality and consistency:
\begin{itemize}
    \item Removal of incorrectly translated or incomplete items.
    \item Correction of formatting inconsistencies in options and rationales.
    \item Verification of alignment between translated answers and rationales with the original source.
\end{itemize}

The final dataset comprises 1,000 high-quality Bangla biomedical Multiple Choice Questions (MCQs) with rationales, offering a more advanced and situational resource compared to the surface-level factual nature of BanglaMedQA. By introducing this dataset, we establish the first advanced Bangla biomedical MCQ corpus for evaluating medical reasoning in a low-resource language. It serves as a critical benchmark for assessing Retrieval Augmented Generation (RAG) and other question-answering strategies in Bangla.

\subsubsection{Data Preprocessing}
Before running experiments, the raw BanglaMedQA dataset and the Bangla MMedBench dataset required several cleaning steps to ensure consistency. First, Unicode normalization was applied to resolve issues with irregular ASCII and full-width characters. Soft and invisible whitespace characters were also normalized to prevent parsing errors. In some cases, the multiple-choice options were embedded directly inside the question text, or stored together under a single key, rather than as separate fields. To address this, a regular expression–based parser was implemented to detect option markers (e.g., “A.”, “B)”, “C:”) and split them into distinct entries for A, B, C, and D. This ensured that every record followed a uniform structure, with a clean question field and exactly four option fields. This cleaned dataset was then used across all subsequent experiments.

\subsection{Retrieval Augmented Generation (RAG)}
Retrieval-Augmented Generation (RAG) is a framework that combines information retrieval with Large Language Models (LLMs) \citep{wu2024retrieval}. Instead of relying solely on the knowledge stored in a language model's parameters, RAG first retrieves relevant documents or context from an external knowledge source (such as a corpus, database, or textbook) \citep{salemi2024evaluating} and then conditions a generative model on this retrieved information to produce accurate and context-aware responses. This approach enhances factual accuracy, allows handling of domain-specific queries, and supports low-resource languages by grounding generation in explicit reference material.

\subsubsection{RAG Retrieval Source: Bangla Biology Textbook}
To construct a reliable retrieval source for our Retrieval-Augmented Generation (RAG) pipelines, we digitized the Higher Secondary level (12th class) Biology textbook written in Bangla, ensuring that the entire content was available in machine-readable form. To ensure the quality and completeness of the digitized text, we followed a two-step process:

\textbf{Step 1: Finding the best OCR system:} Initially, we experimented with Tesseract, an open-source Optical Character Recognition (OCR) engine \citep{smith2007overview, koistinen2017improve}. While Tesseract produced acceptable results for simple text, it struggled with complex Bangla \textit{juktakkhor} (conjunct characters). For example, after processing certain pages, Tesseract often skipped entire paragraphs where such characters appeared, leading to substantial gaps in the extracted text. To address this, we switched to Google Lens OCR \citep{maurya2023googleLens,hegghammer2022ocr}, which yielded significantly cleaner outputs and demonstrated superior handling of complex Bangla character combinations.

\textbf{Step 2: Manual inspection and correction:} Following OCR, the extracted text was carefully verified to correct alignment issues, formatting errors, and occasional missing content. This post-processing step ensured that the corpus retained both structural integrity and linguistic accuracy. The final OCR-processed version of the Higher Secondary level Biology textbook served as the primary knowledge source for RAG-based evaluation, enabling reliable retrieval of domain-specific content directly aligned with Bangla medical Question Answering (QA) tasks.


\subsubsection{Implementation of RAG Variants}

\begin{figure}[t]
    \centering
    \includegraphics[width=\columnwidth]{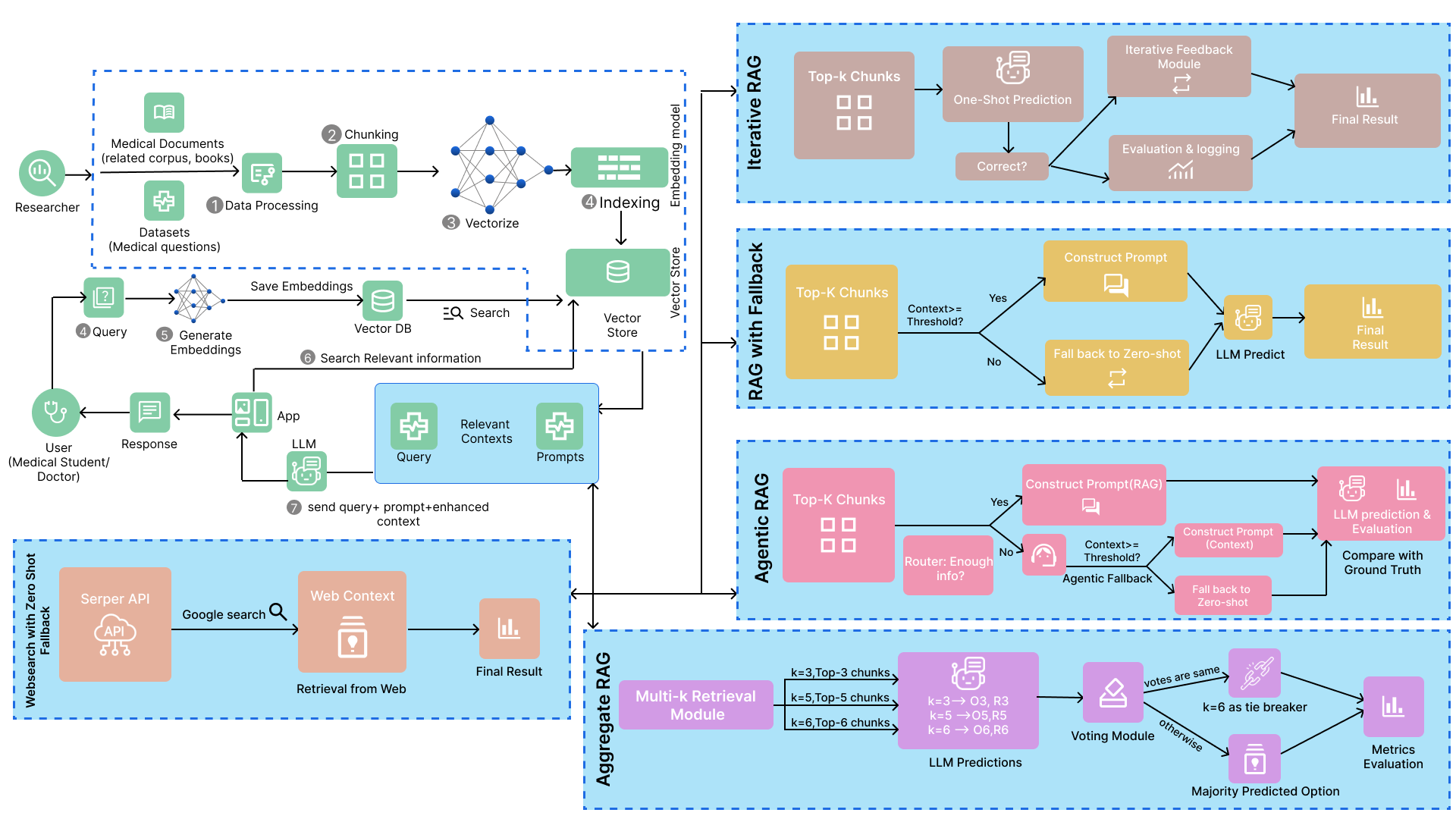} 
    \caption{Methodological workflow of the Bangla RAG framework.}
    \label{fig:bangla-rag-workflow}
\end{figure}

The methodology integrates local retrieval from the textbook with fallback strategies to external web search or zero-shot reasoning. The pipeline shown in \autoref{fig:bangla-rag-workflow} supports multiple strategies, including one-shot reasoning, iterative refinement, and aggregation across different retrieval depths to improve answer accuracy and rationale quality. 



\begin{enumerate}
    \item \textbf{Local RAG:} 
    In traditional Retrieval-Augmented Generation (RAG), a question $q$ with multiple-choice options $O=\{O_A,O_B,O_C,O_D\}$ is input to a Large Language Model (LLM). The model retrieves the top-$k$ passages from a processed Bangla Biology textbook knowledge base $D$ via similarity search:
\[
    C = \text{Retrieve}(q,D,k)
\]
The retrieved context $C$, question $q$, and options $O$ form a prompt for the LLM. If $C$ is sufficient (based on a length threshold), the LLM generates a predicted option $\hat{O} \in O$ and rationale $R$:
\[
    (\hat{O}, R) = \arg\max_a P(a \mid q, O, C)
\]
If $C$ lacks sufficient information, the model outputs:
\[
    (\hat{O}, R) = (\text{N/A}, \;
    \textnormal{\bengali{উত্তর পাওয়া যায়নি}})
\]
where \bengali{"উত্তর পাওয়া যায়নি"} \english means ``The answer was not found'' in Bangla. Experiments showed traditional RAG often returns null responses for questions lacking relevant context, especially for chapter-specific queries. In contrast, zero-shot prompting always produces a prediction $\hat{O}$ and rationale $R$, relying on pretrained knowledge.

\item \textbf{Local RAG with Zero-Shot Fallback:}
To address traditional RAG's null responses when retrieved context is insufficient, we developed a hybrid approach combining retrieval with zero-shot fallback, ensuring all questions receive answers.

The pipeline starts with retrieving top-$k$ passages from knowledge base $D$ for query $q$:
\[
    C = \text{Retrieve}(q, D, k)
\]
The LLM uses $q$, options $O$, and context $C$ to predict answer $\hat{O}$ and rationale $R$:
\[
    (\hat{O}, R) = \arg\max_a P(a \mid q, O, C)
\]
If $C$ is insufficient, instead of outputting \bengali{"উত্তর পাওয়া যায়নি"} \english(``The answer was not found''), the model falls back to zero-shot, using only $q$ and $O$:
\[
    (\hat{O}, R) = \arg\max_a P(a \mid q, O)
\]
This ensures every question has a prediction, balancing precision (with relevant context) and recall (via zero-shot).

\item \textbf{Web Search with Zero-Shot Fallback:}
We developed an Web RAG pipeline for Bangla MCQs, using web retrieval and zero-shot fallback, preserving Comma-Separated Values (CSV) structure with predictions, rationales, and metrics. The steps are shown below:

\textbf{i) Web Retrieval:} The question and its options are used to perform a web search through an API, retrieving up to eight relevant links. The top three results are parsed to extract textual content from key HTML elements such as articles and paragraphs. The extracted information is then summarized into concise Bangla bullet points, forming the web-based contextual input. If the aggregated context exceeds a predefined length threshold, the model generates a structured JSON response.
\[
\begin{aligned}
&\texttt{\{"O": "A|B|C|D",} \\
&\texttt{\ "R": "1-3 line Bangla explanation"\}}
\end{aligned}
\]

\textbf{ii) Zero-Shot Fallback:} If the web context is insufficient, the model resorts to zero-shot reasoning using its pretrained knowledge to generate an answer. HTTP requests, HTML parsing, retry mechanisms, and API delay handling are incorporated to ensure reliability.

This pipeline leverages dynamic web knowledge for Bangla MCQs, with zero-shot fallback ensuring full coverage and compatibility with evaluation.

\item \textbf{Agentic RAG:}
The Agentic RAG pipeline, shown in \autoref{fig:agentic-pipeline}, enhances performance by allowing the LLM to dynamically select the most suitable answering strategy for Bangla MCQs. It adapts between local retrieval, web retrieval, and zero-shot fallback as needed.

\begin{figure}[t]
    \centering
    \includegraphics[width=0.9\columnwidth]{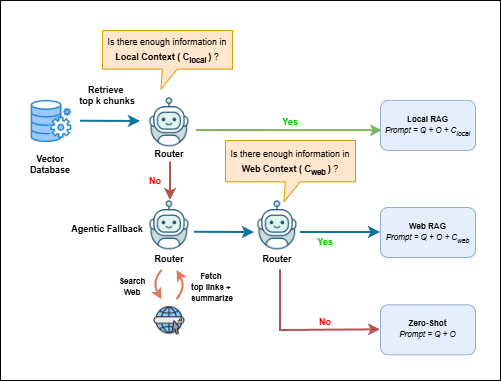} 
    \caption{Agentic RAG pipeline illustration.}
    \label{fig:agentic-pipeline}
\end{figure}

\textbf{i) Local Retrieval:} The query $q$ retrieves top-$k$ passages from textbook $D$:
\[
    C_\text{local} = \text{Retrieve}(q, D, k)
\]

A router checks if $C_\text{local}$ is sufficient:\\

\begin{tcolorbox}[
    colback=gray!10,
    colframe=gray,
    title=নির্দেশনা,
    boxrule=0.7pt,     
    toprule=0.1pt      
]
\begin{bengali}
তুমি একটি রাউটার মডেল। নিচের টেক্সটে কি প্রশ্নের উত্তর দেবার জন্য যথেষ্ট তথ্য আছে? শুধু একটি শব্দে উত্তর দাও:
\end{bengali}
Yes \begin{bengali}অথবা\end{bengali} No।

(Translation: ``You are a router model. Does the text below contain enough information to answer the question? Answer in one word only: Yes or No.'')

\end{tcolorbox}

If Yes and $|C_\text{local}| > \tau_1$, Local RAG generates:
\[
    (\hat{O}, R) = \arg\max_a P(a \mid q, O, C_\text{local})
\]

\textbf{ii) Web Retrieval:} If local context is insufficient, the Serper API fetches up to 8 links, with the top 3 scraped and summarized into Bangla bullet points, forming $C_\text{web}$. If $|C_\text{web}| > \tau_2$, Web RAG generates:
\[
    (\hat{O}, R) = \arg\max_a P(a \mid q, O, C_\text{web})
\]

\textbf{iii) Zero-Shot Fallback:} If both retrievals fail, the model uses zero-shot reasoning:
\[
    (\hat{O}, R) = \arg\max_a P(a \mid q, O)
\]
The routing policy is:
\[
\pi(q) =
\begin{cases} 
\text{Local RAG}, & \text{if Router}(C_\text{local}) = \text{Yes } \wedge \\ & |C_\text{local}| > \tau_1, \\
\text{Web RAG}, & \text{if Router}(C_\text{local}) = \text{No } \wedge \\ & |C_\text{web}| > \tau_2, \\
\text{Zero-Shot}, & \text{otherwise.}
\end{cases}
\]
This ensures flexible knowledge selection and full question coverage.

\item \textbf{Iterative Feedback RAG: }
Iterative Feedback RAG introduces a feedback-driven loop that mimics introspective reasoning as shown in \autoref{fig:rag-pipeline}, improving robustness and accuracy within a single query-response cycle. The pipeline dynamically refines both queries and context through four key phases.

\begin{figure}[t]
    \centering
    \includegraphics[width=0.9\textwidth]{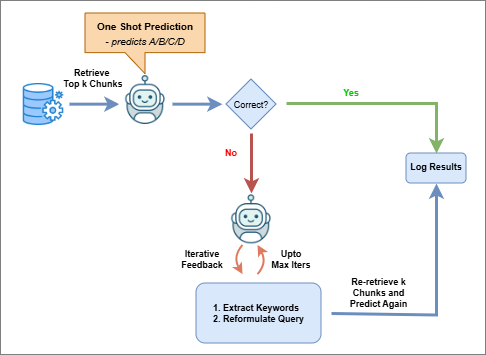} 
    \caption{Iterative Feedback RAG pipeline illustration.}
    \label{fig:rag-pipeline}
\end{figure}

\textbf{i) Context Retrieval:}
The system retrieves the top-$k$ relevant text chunks from a FAISS vector store using similarity search:
\[
C_k = \text{Retrieve}(q, D, k)
\]
Using dense embeddings, the model fetches potentially noisy context, which may require refinement in later steps.

\textbf{ii) One-Shot Answering:}
The LLM generates an initial answer based on the query $q$ and retrieved context $C_k$:
\[
(\hat{O}_0, R_0) = \arg\max P(a \mid q, C_k)
\]
For Multiple-Choice Questions, the model outputs a single character (e.g., ``A''). If the answer is correct, the process terminates; otherwise, it proceeds to the refinement phase.

\textbf{iii) Iterative Refinement:}
If the initial answer is incorrect, the LLM extracts key terms from $q$ and $C_k$ to reformulate the query:
\[
q' = f(q, K)
\]
A new retrieval and prediction are then performed:
\[
(\hat{O}_i, R_i) = \arg\max P(a \mid q', C_k)
\]
Up to two refinement cycles are allowed to enhance context relevance and correct prior reasoning errors, ensuring a balance between efficiency and accuracy.

\textbf{iv) Final Answer and Confidence:}
The system assigns confidence levels based on the outcome of the process:
\begin{itemize}
    \item \textbf{High confidence:} Correct one-shot answer.
    \item \textbf{Medium confidence:} Corrected through refinement.
    \item \textbf{Low confidence:} Incorrect even after refinement.
\end{itemize}
This scoring mechanism enhances interpretability and trust in model outputs.

\begin{figure}[t]
    \centering
    \includegraphics[width=0.9\textwidth]{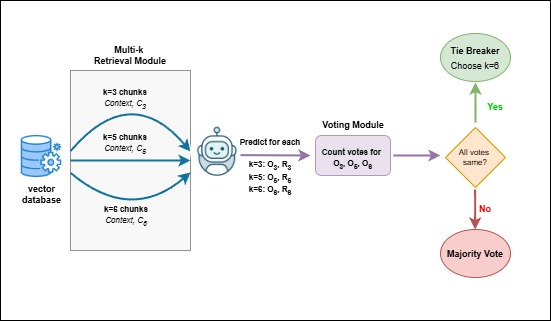} 
    \caption{Aggregate k-values RAG pipeline illustration.}
    \label{fig:aggregate}
\end{figure}

\item \textbf{Aggregate k-values RAG: }
Traditional RAG pipelines retrieve a fixed top-$k$ context, but performance is highly sensitive to the chosen $k$. A small $k$ may miss key information, while a large $k$ can introduce noise. This sensitivity is particularly critical in domains such as zoology, where precise terminology and scattered knowledge are common. Aggregate k-values RAG addresses this limitation by combining predictions from multiple $k$-values as shown in \autoref{fig:aggregate}, thereby improving stability, robustness, and accuracy. The pipeline operates in three stages.

\textbf{i) Context Retrieval:}
For a given query $q$ and knowledge base $D$, the system retrieves context sets for multiple $k$-values:
\[
C_{k_j} = \text{Retrieve}(q, D, k_j), \quad k_j \in \{3, 5, 6\}
\]
Using dense embeddings within a FAISS vector store, different $k$-values serve distinct purposes: $k=3$ prioritizes precision, $k=5$ balances precision and recall, and $k=6$ ensures broad coverage. Together, these retrievals capture complementary pieces of information.

\textbf{ii) LLM Prediction:}
The large language model generates predictions for each context set $C_{k_j}$ as follows:
\[
\hat{O}_{k_j} = \arg\max_{O \in \mathcal{O}} P(O \mid q, C_{k_j})
\]
Each prediction includes a single-character answer (e.g., ``A'') and a brief rationale, produced in a JSON-like format. A temperature of 0.7 is used to balance creativity and determinism, yielding diverse yet coherent responses across different $k$-values.

\textbf{iii) Vote Collection and Aggregation:}
Predictions $\{\hat{O}_{k_j}\}$ are then aggregated through a voting mechanism:
\begin{itemize}
    \item \textbf{Majority:} The most frequent option is selected as the final answer.
    \item \textbf{Unanimous:} Identical predictions across all $k$-values indicate high confidence.
    \item \textbf{Tie:} In case of a tie, the prediction from $k=6$ is chosen, leveraging the broadest context.
\end{itemize}
The final answer is defined as:
\[
\hat{O} = \text{Aggregate}\left(\{\hat{O}_{k_j}\}_{j=1}^m\right), \quad m = 3
\]
The majority rationale corresponding to the winning prediction is selected for consistency and explainability.

\end{enumerate}

\subsection{Comparative Evaluation of MMedBench and BanglaMMedBench}
\begin{figure}[t]
    \centering
    \includegraphics[width=\columnwidth]{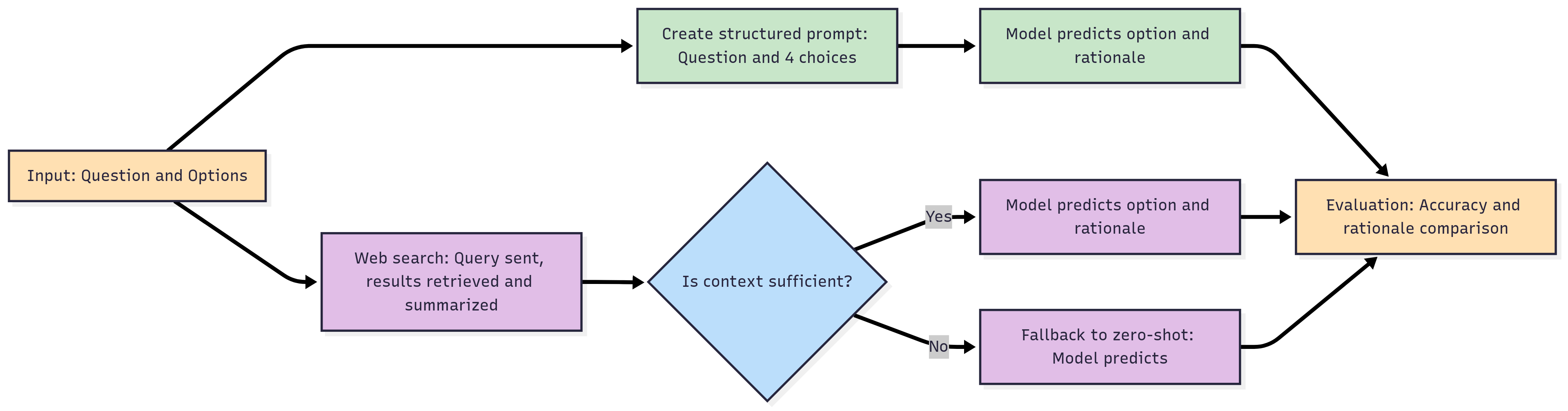} 
    \caption{Zero-shot and web search: MMedBench vs. translated Bangla dataset.}
    \label{fig:mmedbench}
\end{figure}

We evaluated the Zero-Shot and Web Search with Zero-Shot Fallback methods on the MMedBench and BanglaMMedBench datasets, as shown in \autoref{fig:mmedbench}. Traditional RAG was excluded because the scenario-based, patient-specific questions lacked a reliable textbook corpus for retrieval. The experiments compared the English and Bangla versions to assess the impact of translation and the effectiveness of web search integration.

\begin{enumerate}
    \item \textbf{Zero-Shot Analysis:}
The zero-shot evaluation tested the model’s reasoning on both English and Bangla MMedBench datasets without fine-tuning. Each dataset contained Multiple Choice Questions (MCQs) with options \(O_i = \{O_A, O_B, O_C, O_D\}\), ground-truth answers \(a_i\), and, where available, rationales \(r_i\), all preprocessed for consistency. A structured prompt \(P_i = f(q_i, O_i)\) was created for each question \(q_i\) and its options \(O_i\), instructing the model to select a predicted answer \(\hat{a}_i\) and provide a rationale \(\hat{r}_i\) within a 300-second timeout. The evaluation measured accuracy as \(\text{Accuracy} = \frac{1}{N} \sum_{i=1}^{N} \mathbf{1}(\hat{a}_i = a_i)\), while rationales \(\hat{r}_i\) were qualitatively compared to \(r_i\), with results \(R_i = \{\hat{a}_i, \hat{r}_i, a_i, r_i, \text{Accuracy}_i\}\) used to compare language performance.

\item \textbf{Web Search with Fallback to Zero-Shot:}
To enhance the performance of the zero-shot setting, we introduced a pipeline that
augments question answering with external web retrieval while retaining a fallback
to zero-shot when necessary. This design ensures that every question receives an
attempted answer while grounding predictions in external evidence whenever possible.\\

\textbf{i) Web Retrieval:}
Query $q$ uses Serper API for up to 8 links, top 3 scraped into $C_\text{web}$. If $|C_\text{web}| > \tau_2$, it generates:
\[
    (\hat{O}, R) = \arg\max_a P(a \mid q, O, C_\text{web})
\]

\textbf{ii) Zero-Shot Fallback:}
If $|C_\text{web}| \leq \tau_2$ or retrieval fails, it falls back to:
\[
    (\hat{O}, R) = \arg\max_a P(a \mid q, O)
\]
This ensures answers for scenario-based MCQs with limited textbook data.
\end{enumerate}

\subsection{Experimental Setup}

To evaluate the proposed approaches, we employed multiple Large Language Models (LLMs), external APIs, and supporting libraries to construct a robust RAG-based pipeline. The setup integrates both local retrieval from the Bangla biology textbook and fallback strategies to external web search or zero-shot reasoning.

\subsubsection{Large Language Models}
We experimented with a range of models served through the Groq API \citep{groqconsole_orgsettings}, including llama-3.3-70b-versatile \citep{dubey2024llama}, llama-3.1-8b-instant \citep{vavekanand2024llama}, openai/gpt-oss-20b \citep{inuwa2025openai}, and openai-/gpt-oss-120b. The Groq API was chosen for its low-latency inference and ease of switching between different model variants without modifying downstream code.

\subsubsection{Embedding Models}
For Bangla text, we used the BengaliSBERT model, a multilingual variant of Sentence-BERT (SBERT) specifically optimized for semantic similarity in Bangla. SBERT modifies the original BERT architecture by introducing a siamese network structure that enables the generation of semantically meaningful sentence embeddings instead of token-level representations. This allows direct comparison of sentence vectors using cosine similarity, which significantly improves efficiency in semantic search and clustering.

\subsubsection{Vector Database and Chunking}
The entire Bangla biology textbook was normalized, cleaned, and segmented into overlapping chunks of 1,000 characters with an overlap of 200. This chunking ensured that retrieved passages were semantically complete while avoiding information loss at segment boundaries. A FAISS index \citep{douze2025faiss} was built and cached locally for reuse across runs.

\subsubsection{External Web Search}
The Serper API \citep{serper2025} was integrated to perform web retrieval. For each query, up to 8 search results were requested, of which the top 3 links were scraped. Extracted raw text was summarized into a compact context before being passed to the LLM.

\subsubsection{Agentic RAG with Router}
Unlike traditional RAG, which only checks context length, the agentic pipeline introduced a router module that explicitly judged whether retrieved textbook passages contained sufficient information. The router was prompted to answer “Yes” or “No,” and its decision determined whether to proceed with Local RAG or fallback to Web Search. Thresholds were applied to further control routing:

\begin{itemize}
    \item $\tau_1$ = 300 characters for textbook retrieval.
    \item $\tau_2$ = 200 characters for web summaries.
\end{itemize}

If the retrieved or summarized context failed these thresholds, the pipeline defaulted to zero-shot answering.

\subsubsection{Iterative MCQ Evaluation}
To improve answer accuracy, questions that failed the initial one-shot evaluation were reprocessed using an iterative method with 2 iterations. In this approach, the model extracted key keywords or short phrases from the retrieved context in each iteration, refining the query before generating the final answer.

\subsubsection{Aggregated RAG Evaluation}
To enhance prediction robustness, each question was processed with multiple retrieval settings (k = 3, 5, and 6) from the FAISS vector store. The model’s answers from these different retrievals were combined using a voting mechanism, with a tie-breaker favoring the largest retrieval set (k = 6). This aggregation approach reduces errors from incomplete context retrieval and improves both option selection and rationale generation.

\subsubsection{Error Handling and Retry Strategy}
To handle instability from API calls, the system used exponential backoff retries for Groq and Serper requests (up to 3–4 attempts). This ensured smoother execution during large-scale batch runs, preventing failures caused by transient network or server issues.  In addition, detailed logging was implemented to capture error metadata such as request latency, response status, and endpoint behavior. This information was later used to analyze API reliability trends and optimize retry intervals. Combined with local caching of successful responses, the retry strategy improved both resilience and throughput in the overall evaluation pipeline.

\subsubsection{Evaluation Metrics}
To comprehensively evaluate both answer correctness and rationale quality, we utilized a combination of quantitative metrics. Accuracy was used to assess discrete answer prediction performance, while BERTScore, METEOR, ROUGE, and BLEU measured the semantic and lexical similarity between generated and reference rationales.

\begin{enumerate}
    \item \textbf{Accuracy:} Proportion of correctly predicted MCQ answers as in \autoref{eq:accuracy}, indicating how often the model selects the correct option. Higher accuracy reflects better comprehension and retrieval performance.

    \begin{equation}
    \text{Accuracy} = \frac{\text{Number of Correct Predictions}}{\text{Total Number of Predictions}}
    \label{eq:accuracy}
    \end{equation}

    \item \textbf{BERTScore (F1):} Evaluates the semantic similarity between generated and reference rationales using contextual embeddings from a pretrained BERT model. Instead of relying on surface-level n-gram overlap, it compares embeddings token by token, capturing deeper semantic alignment and meaning preservation between two texts \citep{zhang2019bertscore}. The BERTScore (F1) formula is the standard harmonic mean for F1 scores, calculated as:
    \begin{equation}
F_{1} = 2 \times \frac{P \times R}{P + R}
\end{equation}

where \( P \) is the BERT-Precision and \( R \) is the BERT-Recall.

    \item \textbf{METEOR Score:} Considers precision, recall, and synonym matching for textual similarity. Unlike BLEU, METEOR accounts for word stemming and paraphrasing, making it more sensitive to variations in word choice and sentence structure that still convey equivalent meanings \citep{banerjee2005meteor}. The METEOR score is calculated as:

\begin{equation}
F_{mean} = \frac{10PR}{R + 9P}
\end{equation}

\begin{equation}
Penalty = 0.5 \times \left( \frac{\#chunks}{\#unigrams\_matched} \right)^{3}
\end{equation}

\begin{equation}
Score = F_{mean} \times (1 - Penalty)
\end{equation}

    \item \textbf{ROUGE Score:} Measures n-gram overlap and longest common subsequence to assess content coverage and informativeness. ROUGE-1 and ROUGE-2 evaluate unigram and bigram overlaps respectively, while ROUGE-L captures the longest sequence of words in common, reflecting overall content similarity and coherence \citep{lin2004rouge}.\\

    \noindent\textbf{ROUGE-1 and ROUGE-2:}

\begin{equation}
\text{Precision} = \frac{\text{Number of matching n-grams}}{\text{Number of n-grams in candidate summary}}
\end{equation}

\begin{equation}
\text{Recall} = \frac{\text{Number of matching n-grams}}{\text{Number of n-grams in reference summary}}
\end{equation}

\begin{equation}
F_{1} = 2 \times \frac{\text{Precision} \times \text{Recall}}{\text{Precision} + \text{Recall}}
\end{equation}

where, we can change “n-grams” with “unigrams” or “bigrams” depending on whether it’s ROUGE-1 or ROUGE-2.\\

\noindent\textbf{ROUGE-L:}

\begin{equation}
\text{Precision} = \frac{\text{Length of LCS}}{\text{Total number of words in candidate summary}}
\end{equation}

\begin{equation}
\text{Recall} = \frac{\text{Length of LCS}}{\text{Total number of words in reference summary}}
\end{equation}

\begin{equation}
F_{\beta} = (1 + \beta^{2}) \times \frac{\text{Precision} \times \text{Recall}}{(\beta^{2} \times \text{Precision}) + \text{Recall}}
\end{equation}

where, LCS is the Longest Common Subsequence and \(\beta \) is a parameter that balances precision and recall.

    \item \textbf{BLEU Score:} Computes n-gram precision to evaluate fluency and surface-level correspondence with reference text. It rewards outputs that share similar word sequences with reference answers, serving as a traditional benchmark for evaluating the quality and grammatical correctness of generated text \citep{papineni2002bleu}. The BLEU score is computed using the geometric mean of modified n-gram precisions and a brevity penalty factor.

\begin{equation}
BP =
\begin{cases}
1, & \text{if } c > r \\
e^{(1 - r/c)}, & \text{if } c \leq r
\end{cases}
\end{equation}

\begin{equation}
BLEU = BP \times \exp \left( \sum_{n=1}^{N} w_n \log p_n \right)
\end{equation}

where \( p_n \) is the modified n-gram precision, \( w_n \) are positive weights such that \( \sum_{n=1}^{N} w_n = 1 \),  
\( c \) is the length of the candidate translation, and \( r \) is the reference length.
\end{enumerate}

\textbf{Rationale for Metric Selection:} These metrics were chosen to provide a holistic evaluation of both answer accuracy and rationale quality. Accuracy serves as a direct measure of the model’s correctness in selecting the right option for MCQs, while BERTScore and METEOR capture semantic and linguistic alignment, assessing whether the model’s reasoning conveys the intended meaning even if exact words differ. ROUGE metrics evaluate the extent to which the generated rationales cover relevant content from the reference, including structural and sequence similarities. BLEU scores complement this by assessing surface-level fluency and n-gram precision, ensuring that generated explanations are coherent and readable.  

By combining these metrics, the evaluation captures multiple dimensions of performance: factual correctness, semantic fidelity, content coverage, and linguistic quality. This is particularly important for assessing models across both English and Bangla datasets, where the goal is to measure not only accuracy but also the quality and clarity of the generated rationales.

\begin{table*}[htbp]
\centering
\caption{Performance Metrics on BanglaMedQA Across Different RAG Configurations. The best results for each metric are shown in bold.}
\resizebox{\textwidth}{!}{%
\setlength{\tabcolsep}{1pt}
\scriptsize
\renewcommand{\arraystretch}{0.9}
\begin{tabular}{lcccccccc}
\toprule
\textbf{Model} & \textbf{Accuracy} & \textbf{BERT} & \textbf{METEOR} & \textbf{ROUGE-1} & \textbf{ROUGE-2} & \textbf{ROUGE-L} & \textbf{BLEU-1} & \textbf{BLEU-2} \\
\midrule
\multicolumn{9}{c}{\textbf{Zero-Shot}} \\
\midrule
llama-3.3-70b-versatile & 73.04\% & 0.7557 & 0.1426 & 0.2063 & 0.0714 & \textbf{0.1853} & 0.1414 & 0.0804 \\
llama-3.1-8b-instant & 42.15\% & 0.7244 & 0.1149 & 0.1488 & 0.0374 & 0.1299 & 0.1163 & 0.0547 \\
openai/gpt-oss-120b & 82.49\% & 0.7300 & 0.1024 & 0.1506 & 0.0325 & 0.1318 & 0.1143 & 0.0509 \\
openai/gpt-oss-20b & 82.60\% & 0.7212 & 0.1048 & 0.1544 & 0.0363 & 0.1362 & 0.1143 & 0.0525 \\
\midrule
\multicolumn{9}{c}{\textbf{Local RAG}} \\
\midrule
llama-3.3-70b-versatile & 82.70\% & 0.7638 & 0.1444 & 0.0603 & 0.0289 & 0.0586 & 0.1294 & 0.0788 \\
llama-3.1-8b-instant & 65.79\% & 0.7422 & 0.1451 & 0.0465 & 0.0200 & 0.0457 & 0.1403 & 0.0782 \\
openai/gpt-oss-120b & 86.32\% & 0.7347 & 0.1148 & 0.0825 & 0.0284 & 0.0803 & 0.1215 & 0.0590 \\
openai/gpt-oss-20b & 83.50\% & 0.7329 & 0.1220 & 0.0685 & 0.0186 & 0.0675 & 0.1270 & 0.0636 \\
\midrule
\multicolumn{9}{c}{\textbf{Local RAG + Zero-Shot Fallback}} \\
\midrule
llama-3.3-70b-versatile & 82.90\% & 0.7657 & 0.1482 & 0.0560 & 0.0295 & 0.0556 & 0.1366 & 0.0821 \\
llama-3.1-8b-instant & 66.60\% & 0.7333 & \textbf{0.1517} & 0.0562 & 0.0231 & 0.0541 & 0.1424 & 0.0776 \\
openai/gpt-oss-120b & 87.22\% & 0.7078 & 0.1137 & 0.0796 & 0.0273 & 0.0774 & 0.1234 & 0.0565 \\
openai/gpt-oss-20b & 85.81\% & 0.6984 & 0.1193 & 0.0726 & 0.0208 & 0.0700 & 0.1265 & 0.0609 \\
\midrule
\multicolumn{9}{c}{\textbf{Web Search + Zero-Shot Fallback}} \\
\midrule
llama-3.3-70b-versatile & 87.83\% & 0.7564 & 0.1469 & 0.0468 & 0.0162 & 0.0458 & 0.1422 & 0.0783 \\
llama-3.1-8b-instant & 58.35\% & 0.7141 & 0.1473 & 0.0505 & 0.0222 & 0.0497 & 0.1316 & 0.0679 \\
openai/gpt-oss-120b & 88.83\% & 0.7126 & 0.1048 & 0.0827 & 0.0226 & 0.0805 & 0.1151 & 0.0516 \\
openai/gpt-oss-20b & 87.02\% & 0.7122 & 0.1140 & 0.0657 & 0.0199 & 0.0642 & 0.1197 & 0.0571 \\
\midrule
\multicolumn{9}{c}{\textbf{Agentic RAG (Local + Web) + Zero-Shot Fallback}} \\
\midrule
llama-3.3-70b-versatile & 86.62\% & 0.7593 & 0.1476 & 0.0494 & 0.0208 & 0.0483 & 0.1429 & \textbf{0.0841} \\
llama-3.1-8b-instant & 63.38\% & 0.7238 & 0.1459 & 0.0520 & 0.0177 & 0.0504 & 0.1396 & 0.0749 \\
openai/gpt-oss-120b & \textbf{89.54\%} & 0.7274 & 0.1136 & 0.0897 & 0.0271 & 0.0870 & 0.1228 & 0.0565 \\
openai/gpt-oss-20b & 88.83\% & 0.7353 & 0.1253 & 0.0763 & 0.0228 & 0.0746 & 0.1339 & 0.0669 \\
\midrule
\multicolumn{9}{c}{\textbf{Iterative Feedback RAG}} \\
\midrule
llama-3.3-70b-versatile & 78.07\% & 0.7431 & 0.1424 & 0.0643 & 0.0174 & 0.1462 & 0.1129 & 0.0754 \\
llama-3.1-8b-instant & 49.20\% & \textbf{0.7982} & 0.1352 & 0.0432 & 0.0343 & 0.0563 & \textbf{0.1432} & 0.0723 \\
openai/gpt-oss-120b & 88.73\% & 0.7537 & 0.1348 & 0.0225 & 0.0255 & 0.0801 & 0.1115 & 0.0591 \\
openai/gpt-oss-20b & 87.42\% & 0.7231 & 0.1230 & 0.0655 & 0.0176 & 0.0635 & 0.1170 & 0.0646 \\
\midrule
\multicolumn{9}{c}{\textbf{Aggregate k-values RAG}} \\
\midrule
llama-3.3-70b-versatile & 82.34\% & 0.6526 & 0.1211 & 0.0621 & 0.0254 & 0.0532 & 0.1424 & 0.0634 \\
llama-3.1-8b-instant & 51.81\% & 0.6321 & 0.1452 & \textbf{0.2314} & \textbf{0.0732} & 0.1642 & 0.1113 & 0.0697 \\
openai/gpt-oss-120b & 84.51\% & 0.6925 & 0.1059 & 0.1412 & 0.0391 & 0.1227 & 0.1124 & 0.0579 \\
openai/gpt-oss-20b & 83.95\% & 0.6688 & 0.0938 & 0.1221 & 0.0310 & 0.1074 & 0.0951 & 0.0453 \\
\bottomrule
\end{tabular}
}
\label{tab:all}
\end{table*}

\section{Results and Discussion}\label{sec4}
\subsection{Presenting the Results}

Across all methods, as shown in \autoref{tab:all}, Zero-Shot serves as the baseline with accuracies from 42.15\% (llama-3.1-8b-instant) to 82.60\% (openai/gpt-oss-20b). Traditional RAG improves to 65.79\%--86.32\%, while RAG with Zero-Shot Fallback reaches 87.22\%, and Iterative Feedback RAG offers modest gains but underperforms advanced methods. Agentic RAG achieves the highest accuracy at 89.54\% (openai/gpt-oss-120b), showcasing its dynamic retrieval strategy, followed by Aggregate k-values RAG at 84.51\% (openai/gpt-oss-120b). Larger models like llama-3.3-70b-versatile outperform smaller variants (e.g., llama-3.1-8b-instant), though gains diminish beyond openai/gpt-oss-20b.

Agentic RAG also excels in rationale quality, with superior ROUGE-1, ROUGE-L, and BLEU scores, indicating strong linguistic and semantic coherence. Zero-Shot and Iterative Feedback RAG show solid BERTScore and METEOR values, preserving semantic meaning, while Traditional and Fallback RAG maintain balanced performance. Aggregate k-values RAG remains consistent despite retrieval noise.

In summary, Agentic RAG with Zero-Shot Fallback offers the best combination of accuracy and rationale quality, making it the most effective approach for BanglaMedQA.

\begin{table*}[t]
\centering
\caption{Accuracy Values for MMedBench Dataset and Its Bangla Translation}
\setlength{\tabcolsep}{4pt}
\scriptsize
\renewcommand{\arraystretch}{1.15}
\begin{tabular}{lcccc}
\toprule
\multirow{2}{*}{\textbf{Model}} & 
\multicolumn{2}{c}{\textbf{Bangla Dataset}} & 
\multicolumn{2}{c}{\textbf{English Dataset}} \\
\cmidrule(lr){2-3} \cmidrule(lr){4-5}
 & \textbf{Zero-Shot} & \textbf{Web Search} & \textbf{Zero-Shot} & \textbf{Web Search} \\
\midrule
llama-3.3-70b-versatile & 62.08\% & 60.00\% & 88.90\% & 83.86\% \\
openai/gpt-oss-120b & 90.59\% & 82.97\% & 92.47\% & 89.90\% \\
\bottomrule
\end{tabular}
\label{tab:mmedbench}
\end{table*}

\autoref{tab:mmedbench} presents accuracy values for the MMedBench dataset and its Bangla translation across models and settings. For Bangla, openai/gpt-oss-120b achieved 90.59\%, outperforming llama-3.3-70b-versatile at 62.08\%. With Web Search, accuracies dropped to 82.97\% and 60.00\%, respectively. On the English dataset, openai/gpt-oss-120b reached 92.47\% (Zero-Shot) and 89.90\% (Web Search), while llama-3.3-70b-versatile scored 88.90\% and 83.86\%. Zero-Shot performs better on English, while Web Search often reduces accuracy, especially for BanglaMMedBench, due to scenario-based questions where generic web content offers limited value. 

\subsection{Interpreting the Results}

Traditional RAG improved performance over Zero-Shot by grounding answers in textbook content, as shown in \autoref{tab:all}, confirming that external knowledge enhances factual accuracy. However, gains varied across models due to differences in embedding quality, Llama models showed larger jumps than GPT-oss, which already performed well in Zero-Shot. The fallback strategy brought minor (1–2\%) improvements, mainly by avoiding unanswered questions, though it sometimes retained irrelevant long passages. 

Agentic RAG achieved the best overall results, especially with larger models such as \texttt{openai/gpt-oss-120b}. Its routing mechanism reduced reliance on context length, enhancing robustness, though occasional errors arose from imperfect routing and noisy web data. Iterative Feedback RAG improved over Zero-Shot but inconsistently; for instance, it reached 78.07\% accuracy for \texttt{llama-3.3-70b-versatile} versus 88.73\% for \texttt{openai/gpt-oss-120b}, indicating sensitivity to model capacity and initial retrieval quality. Errors could propagate during iterations, limiting reliability. Aggregate k-values RAG demonstrated stable performance, just below Agentic RAG, by combining predictions across multiple retrieval sizes. Its accuracies, 82.34\% (\texttt{llama-3.3-70b-versatile}) and 84.51\% (\texttt{openai/gpt-oss-120b}), indicate that aggregating diverse contexts balances precision and coverage, reducing missed information.

On the MMedBench dataset and its Bangla translation, as shown in \autoref{tab:mmedbench}, Zero-Shot and Web Search methods performed better on the English version, reflecting the models’ English-centric pretraining. In the Bangla version, \texttt{openai/gpt-oss-120b} surpassed \texttt{llama-3.3-70b-versatile}, suggesting that model scale can help reduce cross-lingual gaps. However, Web Search sometimes decreased accuracy, as generic web content rarely aligned with the scenario-based, clinical nature of MMedBench questions. This shows that web grounding does not always guarantee performance improvement across domains and languages.

\section{Conclusion}\label{sec5}

This study introduced BanglaMedQA and BanglaMMedBench, two Bangla biomedical Multiple Choice Question (MCQ) datasets. Several Retrieval-Augmented Generation (RAG) strategies were evaluated for question answering in a low-resource setting. Using a Bangla medical textbook corpus for retrieval, the Agentic RAG approach achieved the highest accuracy of 89.54\% with openai/gpt-oss-120b. Iterative Feedback RAG also showed consistent and interpretable results across models. LLaMA models benefited more from retrieval-based augmentation, while web search retrieval sometimes reduced accuracy due to content mismatch. Despite limitations in dataset size, translation quality, and computational resources, this work provides the first comprehensive benchmark for Bangla biomedical QA and demonstrates the potential of retrieval-based reasoning in advancing multilingual medical artificial intelligence.

\section*{CRediT authorship contribution statement}

\textbf{Sadia Sultana:} Conceptualization, Data curation, Methodology, Formal analysis. 
\textbf{Saiyma Sittul Muna:} Conceptualization, Data curation, Investigation, Methodology. 
\textbf{Mosammat Zannatul Samarukh:} Conceptualization, Data curation, Visualization, Formal analysis. 
\textbf{Ajwad Abrar:} Writing – review \& editing, Supervision, Project administration. 
\textbf{Tareque Mohmud Chowdhury:} Supervision, Validation, Funding acquisition. 

\section*{Data Availability Statement}
All data is available on a public Hugging Face repository: \url{https://huggingface.co/datasets/ajwad-abrar/BanglaMedQA}.

\section*{Declaration of competing interest}
The authors declare that they have no known competing financial interests or personal relationships that could have appeared to influence the work reported in this paper.

 \bibliographystyle{elsarticle-harv} 
 \bibliography{references}

\end{document}